\documentclass[10pt,twocolumn,letterpaper]{article}

\usepackage{iccv}
\usepackage{times}
\usepackage{epsfig}
\usepackage{graphicx}
\usepackage{amsmath}
\usepackage{amssymb}
\usepackage{bm}
\usepackage{algorithm}
\usepackage{algorithmic}
\usepackage{authblk}

\usepackage[breaklinks=true,bookmarks=false]{hyperref}

\iccvfinalcopy 

\def\iccvPaperID{****} 

\begin{document}

\title{CVAE-GAN: Fine-Grained Image Generation through Asymmetric Training}

\author{Jianmin Bao$^{1}$, \quad Dong Chen$^{2}$, \quad Fang Wen$^{2}$, \quad Houqiang Li$^{1}$, \quad Gang Hua$^{2}$\\
$^{1}$University of Science and Technology of China \qquad $^{2}$Microsoft Research\\
{\tt\small jmbao@mail.ustc.edu.cn}  \quad{\tt\small \{doch, fangwen, ganghua\}@microsoft.com}  \quad{\tt\small lihq@ustc.edu.cn} }

\maketitle

\hyphenation{in-suf-fi-cient}

\begin{abstract}

We present variational generative adversarial networks, a general learning framework that combines a variational auto-encoder with a generative adversarial network, for synthesizing images in fine-grained categories, such as faces of a specific person or objects in a category. Our approach models an image as a composition of label and latent attributes in a probabilistic model. By varying the fine-grained category label fed into the resulting generative model, we can generate images in a specific category with randomly drawn values on a latent attribute vector. Our approach has two novel aspects. First, we adopt a cross entropy loss for the discriminative and classifier network, but a mean discrepancy objective for the generative network. This kind of asymmetric loss function makes the GAN training more stable. Second, we adopt an encoder network to learn the relationship between the latent space and the real image space, and use pairwise feature matching to keep the structure of generated images. We experiment with natural images of faces, flowers, and birds, and demonstrate that the proposed models are capable of generating realistic and diverse samples with fine-grained category labels. We further show that our models can be applied to other tasks, such as image inpainting, super-resolution, and data augmentation for training better face recognition models.

\end{abstract}

\section{Introduction}
Building effective generative models of natural images is one of the key problems in computer vision. It aims to generate diverse realistic images by varying some latent parameters according to the underlying natural image distributions. Therefore, a desired generative model is necessitated to capture the underlying data distribution. This is often a very difficult task, since a collection of image samples may lie on a very complex manifold. Nevertheless, recent advances in deep convolutional neural networks have spawned a series of deep generative models~\cite{larochelle2011neural,kingma2013auto,goodfellow2014generative,rezende2014stochastic,radford2015unsupervised,sohn2015learning,larsen2015autoencoding,denton2015deep,salimans2016improved,dosovitskiy2016learning} that have made tremendous progress, largely due to the capability of deep networks in learning representations.

\begin{figure}
  \centering
  \includegraphics[width=\columnwidth]{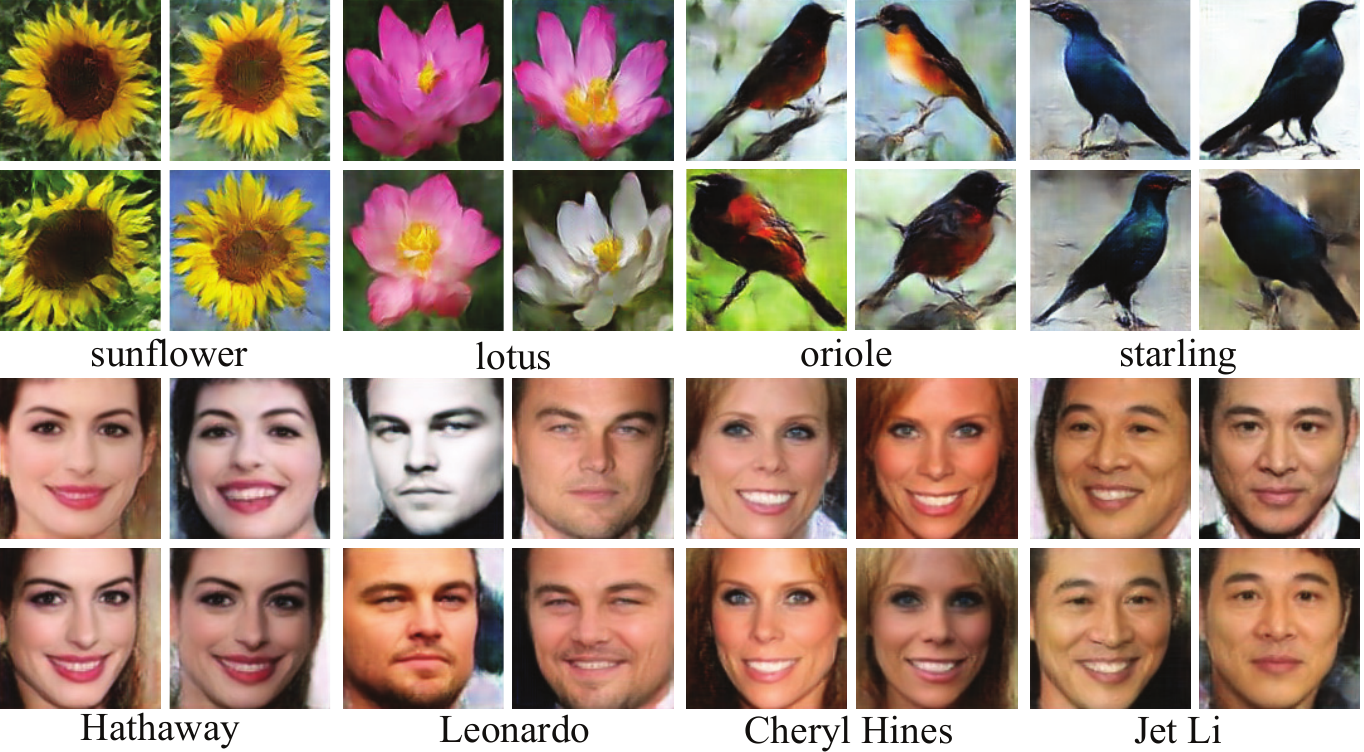}\\
  \caption{\;\textbf{Synthesized\; images}\; using our CVAE-GAN model at high resolution (128$\times$128) for different classes. The generated samples are realistic and diverse within a class.}\label{fig:intro}
  \vspace{-0.4cm}
\end{figure}

Building on top of the success of these recent works, we want to go one step further to generate images of fine-grained object categories. For example, we want to be able to synthesize images for a specific identity (Figure~\ref{fig:intro}), or produce a new image of a specified species of flowers or birds, and so on. Inspired by CVAE~\cite{sohn2015learning} and VAE/GAN~\cite{larsen2015autoencoding}, we propose a general learning framework that combines a variational auto-encoder with a generative adversarial network under a conditioned generative process to tackle this problem.

However, we find this na\"ive combination insufficient in practice. The results from VAE are usually blurry. The discriminator can easily classify them as ``fake", even though they sometimes look remarkably good for face images, the gradient vanishing problem still exists. Thus, the generated images are very similar to the results from using VAE alone.

In this paper, we propose a new objective for the generator. Instead of using the same cross entropy loss as the discriminator network, the new objective requires the generator to generate data that minimize the $\ell_2$ distance of the mean feature to the real data. For multi-class image generation, the generated samples of one category also need to match the average feature of real data of that category, since the feature distance and the separability are positively correlated. It solves the gradient vanishing problem to a certain extent. This kind of asymmetric loss function can partially help prevent the mode collapse problem that all outputs moving toward a single point, making the training of GAN more stable.

Although using mean feature matching will reduce the chance of mode collapse, it does not completely solve the problem. Once mode collapse occurs, the gradient descent is unable to separate identical outputs. To keep the diversity of generated samples, we take advantage of the combination of VAE and GAN. We use an encoder network to map the real image to the latent vector. Then the generator is required to reconstruct the raw pixels and match the feature of original images with a given latent vector. In this way, we explicitly set up the relationship between the latent space and real image space. Because of the existence of these anchor points, the generator is enforced to emit diverse samples. Moreover, the pixel reconstruction loss is also helpful for maintaining the structure, such as a straight line or a facial structure in an image.

As shown in Figure~\ref{fig:VAE_GAN} (g), our framework consists of four parts: 1) The encoder network $E$, which maps the data sample $\bm{x}$ to a latent representation $\bm{z}$. 2) The generative network $G$, which generates image $\bm{x}'$ given a latent vector. 3) The discriminative network $D$, which distinguishes real/fake images.  4) The classifier network $C$, which measures the class probability of the data. These four parts are seamlessly cascaded together, and the whole pipeline is trained end-to-end. We call our approach CVAE-GAN.

Once the CVAE-GAN is trained, it can be used in different applications, {\em e.g.},  image generation, image inpainting, and attributes morphing. Our approach estimates a good representation of the input image, and the generated image appears to be more realistic. We show that it outperforms CVAE, CGAN, and other state-of-the-art methods. Compared with GAN, the proposed framework is much easier to train and converges faster and more stable in the training stage. In our experiments, we further show that the images synthesized from our models can be applied to other tasks, such as data augmentation for training better face recognition models.

\section{Related work}

Conventional wisdom and early research of generative models, including Principle Component Analysis (PCA)~\cite{turk1991face}, Independent Component Analysis (ICA)~\cite{hyvarinen2004independent},and the Gaussian Mixture Model (GMM)~\cite{xu1996convergence,permuter2003gaussian,theis2012mixtures}, all assume a simple formation of data. They have difficulty modeling complex patterns of irregular distributions. Later works, such as the Hidden Markov Model (HMM)~\cite{starner1997real}, Markov Random Field (MRF)~\cite{mnih2010generating}, and restricted Boltzmann machines (RBMs)~\cite{hinton2006reducing,salakhutdinov2009deep}, discriminatively train generative models~\cite{tu2007learning}, limiting their results on texture patches, digital numbers or well aligned faces, due to a lack of effective feature representations.

There have been many recent developments of deep generative models~\cite{larochelle2011neural,kingma2013auto,goodfellow2014generative,rezende2014stochastic,radford2015unsupervised,larsen2015autoencoding,denton2015deep,salimans2016improved,dosovitskiy2016learning}. Since deep hierarchical architectures allow them to capture complex structures in the data, all these methods show promising results in generating natural images that are far more realistic than conventional generative models. Among them are three main themes: Variational Auto-encoder (VAE)~\cite{kingma2013auto,rezende2014stochastic}, Generative Adversarial Network (GAN)~\cite{goodfellow2014generative,radford2015unsupervised,salimans2016improved}, and Autoregression~\cite{larochelle2011neural}.

\begin{figure}
  \centering
  \includegraphics[width=0.9\columnwidth]{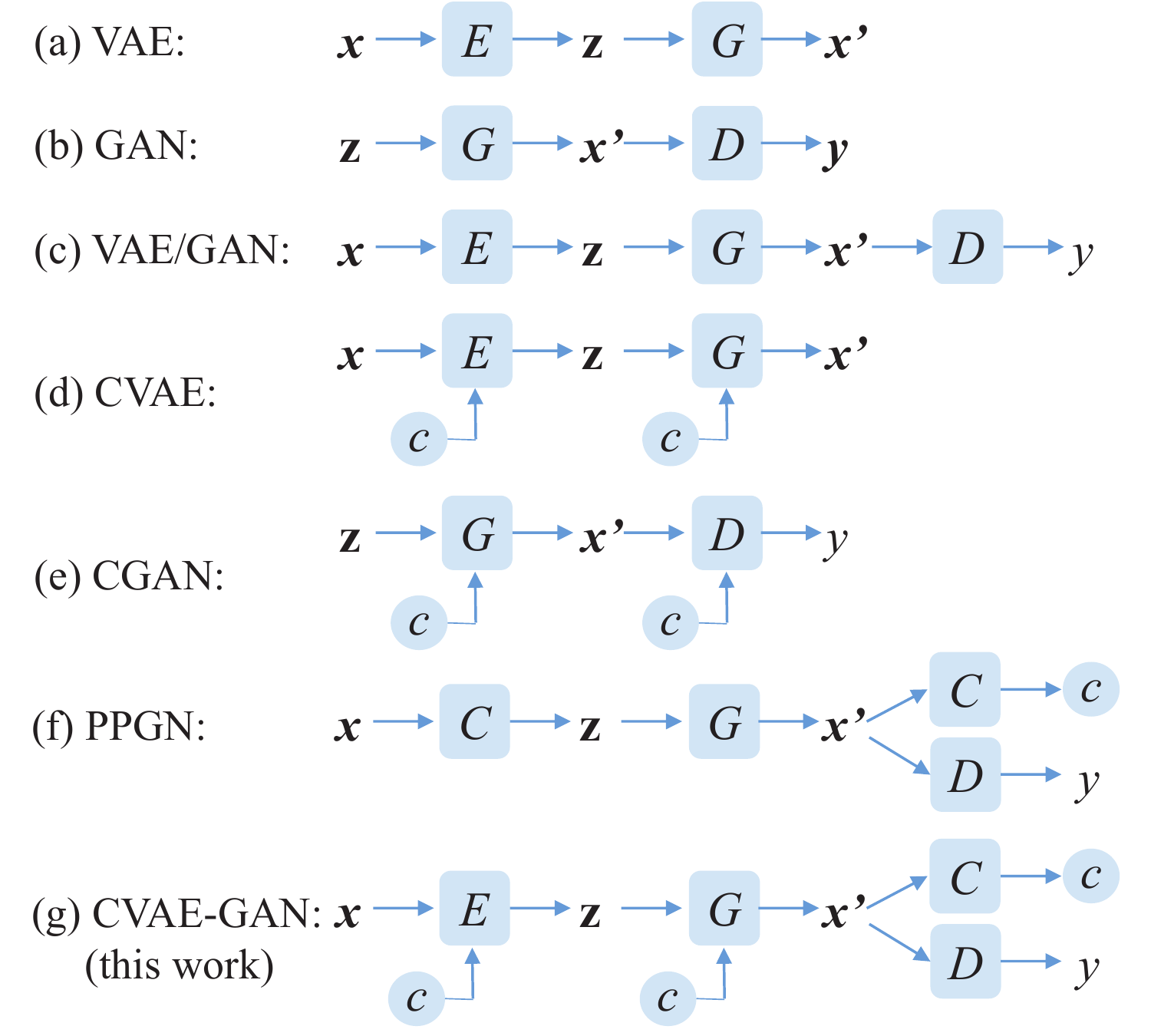}\\
  \caption{Illustration of the structure of VAE~\cite{kingma2013auto,rezende2014stochastic}, GAN~\cite{goodfellow2014generative}, VAE/GAN~\cite{larsen2015autoencoding}, CVAE~\cite{sohn2015learning}, CGAN~\cite{mirza2014conditional}, PPGN~\cite{nguyen2016plug} and the proposed CVAE-GAN. Where $\bm{x}$ and $\bm{x}'$ are input and generated image. $E$,$G$,$C$,$D$ are encoder, generative, classification, and discriminative network, respectively. $\bm{z}$ is the latent vector. $y$ is a binary output which represents real/synthesized image. $c$ is the condition, such as attribute or class label.}
  \label{fig:VAE_GAN}
  \vspace{-0.5cm}
\end{figure}

VAE~\cite{kingma2013auto,rezende2014stochastic} pairs a differentiable encoder network with a decoder/generative network. A disadvantage of VAE is that, because of the injected noise and imperfect element-wise measures such as the squared error, the generated samples are often blurry.

Generative Adversarial Network (GAN)~\cite{goodfellow2014generative,radford2015unsupervised,salimans2016improved} is another popular generative model. It simultaneously trains two models: a generative model  to synthesize samples, and a discriminative model to differentiate between natural and synthesized samples. However, the GAN model is hard to converge in the training stage and the samples generated from GAN are often far from natural. Recently, many works have tried to improve the quality of the generated samples. For example, the Wasserstein GAN (WGAN)~\cite{arjovsky2017wasserstein} uses Earth Mover Distance as an objective for training GANs, and McGAN~\cite{mroueh2017mcgan} uses mean and covariance feature matching. They need to limit the range of the parameters of the discriminator which will decrease discriminative power. Loss-Sensitive GAN~\cite{qi2017loss} learns a loss function which can quantify the quality of generated samples and uses this loss function to generate high-quality images. There are also methods which tried to combine GAN and VAE, {\em e.g.}, VAE/GAN~\cite{larsen2015autoencoding} and adversarial autoencoders~\cite{makhzani2015adversarial}. They are closely related to and partly inspired our work.

VAEs and GANs can also be trained to conduct conditional generation, {\em e.g.},  CVAE~\cite{sohn2015learning} and CGAN~\cite{mirza2014conditional}. By introducing additional conditionality, they can handle probabilistic one-to-many mapping problems. Recently there have been many interesting works based on CVAE and CGAN, including conditional face generation~\cite{gauthier2014conditional}, Attribute2Image~\cite{yan2015attribute2image}, text to image synthesis~\cite{reed2016generative}, forecasting from static images~\cite{walker2016uncertain}, and conditional image synthesis~\cite{odena2016conditional}. All of them achieve impressive results.

Generative ConvNet~\cite{xie2016theory}, demonstrates that a generative model can be derived from the commonly used discriminative ConvNet. Dosovitskiy et al. ~\cite{dosovitskiy2016generating} and Nguyen et al.~\cite{nguyen2016synthesizing}  introduce a method that generates high quality images from features extracted from a trained classification model. PPGN~\cite{nguyen2016plug} performs exceptionally well in generating samples by using a gradient ascent and prior to the latent space of a generator.

Autoregression~\cite{larochelle2011neural} follows a different idea. It uses autoregressive connections to model images pixel by pixel. Its two variants, PixelRNN~\cite{van2016pixel} and PixelCNN~\cite{oord2016conditional}, also produce excellent samples.

Our model differs from all these models. As illustrated in Figure~\ref{fig:VAE_GAN}, we compare the structure of the proposed CVAE-GAN with all these models. Besides the difference in the structure, more importantly, we take advantages of both statistic and pairwise feature matching to make the training process converge faster and more stable.

\section{Our Formulation of CVAE-GAN}
\label{sec:our_formulation}

\begin{figure}[t]
  \centering
  \includegraphics[width=1\columnwidth]{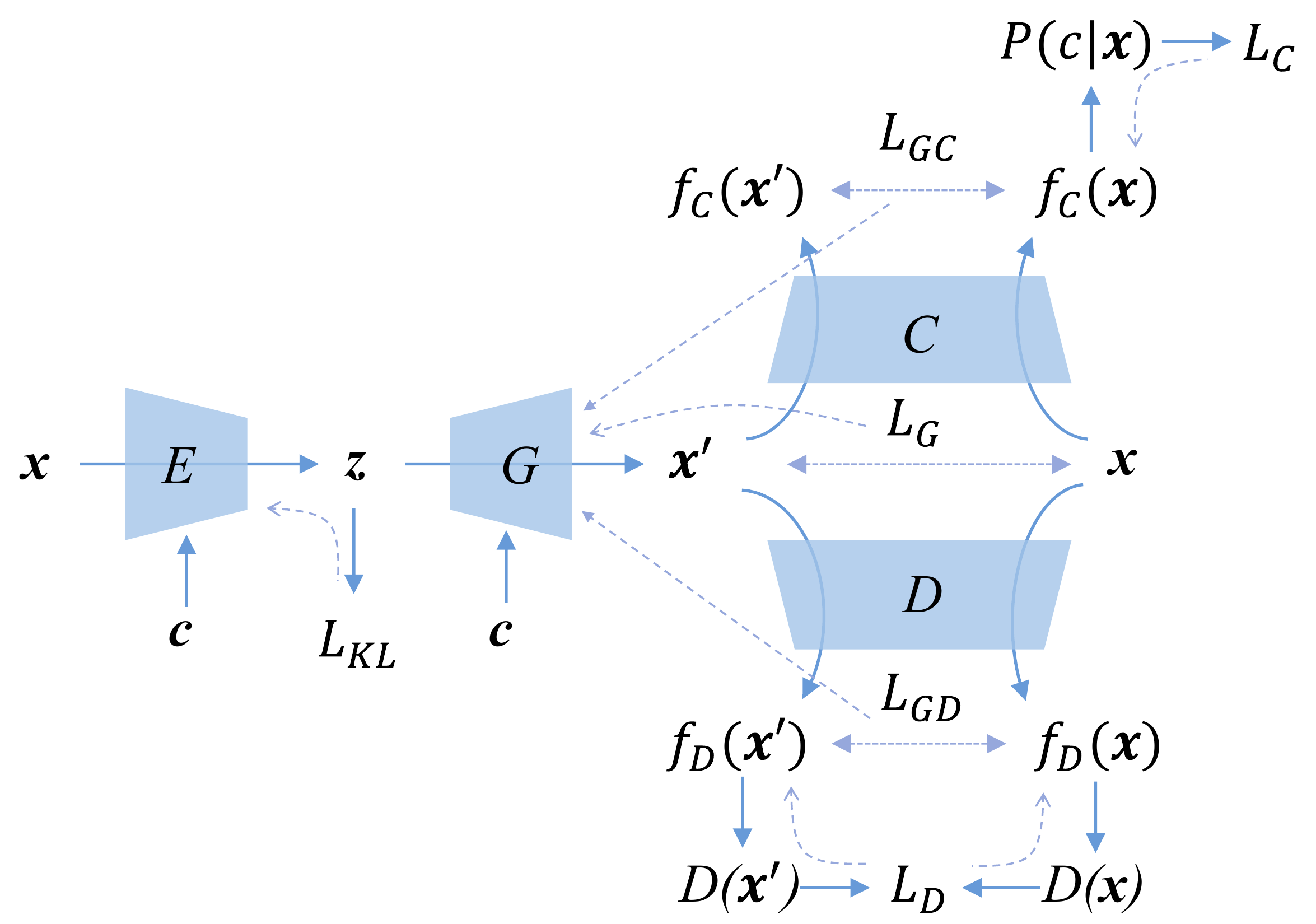}\\
  \caption{Illustration of our network structure. Our model contains four parts: 1) The encoder network $E$; 2) The generative network $G$; 3) The classification network $C$; and 4) The discriminative network $D$. Please refer to Section~\ref{sec:our_formulation} for details.}\label{fig:pipeline}
  \vspace{-0.5cm}
\end{figure}

In this section, we introduce the proposed CVAE-GAN networks. As shown in Figure~\ref{fig:pipeline}, our proposed method contains four parts: 1) the encoder network $E$; 2) the generative network $G$; 3) the discriminative network $D$;  and 4) the classification network $C$.

The function of networks $E$ and $G$ is the same as that in conditional variational auto-encoder (CVAE)~\cite{sohn2015learning}. The encoder network $E$ maps the data sample $\bm{x}$ to a latent representation $\bm{z}$ through a learned distribution $P(\bm{z}|\bm{x},c)$, where $c$ is the category of the data. The generative network $G$ generates image $\bm{x}'$ by sampling from a learned distribution $P(\bm{x}|\bm{z},c)$. The function of network $G$ and $D$ is the same as that in the generative adversarial network (GAN)~\cite{goodfellow2014generative}. The network $G$ tries to learn the real data distribution by the gradients given by the discriminative network $D$ which learns to distinguish between ``real" and ``fake" samples. The function of network $C$ is to measure the posterior $P(c|\bm{x})$.

However, the na\"ive combination of VAE and GAN is insufficient. Recent work ~\cite{arjovsky2017towards} shows that the training of GAN will suffer from a gradient vanishing or instability problem with network $G$. Therefore, we only keep the training process of networks $E$, $D$, and $C$ the same as the original VAE~\cite{kingma2013auto} and GAN~\cite{goodfellow2014generative}, and propose a new mean feature matching objective for the generative network $G$ to improve the stability of the original GAN.

Even with the mean feature matching objective, there is still some risk of mode collapse. So we use the encoder network $E$ and the generative network $G$ to obtain a mapping from real samples $\bm{x}$ to the synthesized samples $\bm{x'}$. By using the pixel-wise $\ell_2$ loss and pair-wise feature matching, the generative model is enforced to emit diverse samples and generate structure-preserving samples.

In the following sections, we begin by describing the method of mean feature matching based GAN (Section~\ref{sec:mean_feature_matching_based GAN}). Then we show that the mean feature matching can also be used in conditional image generation tasks (Section~\ref{sec:mean_feature_matching_based CGAN}). After that, we introduce pair-wise feature matching by using an additional encoder network (Section~\ref{sec:pairwise_feature_matching_based_CVAEGAN}). Finally, we analyse the objective of the proposed method and provide the implementation details in the training pipeline (Section~\ref{sec:objective_of_CVAE_GAN}).

\subsection{Mean feature matching based GAN}
\label{sec:mean_feature_matching_based GAN}
In traditional GANs, the generator $G$ and a discriminator $D$ compete in a two-player minimax game. The discriminator tries to distinguish real training data from synthesized data; and the generator tries to fool the discriminator. Concretely, the network $D$ tries to minimize the loss function
\begin{equation}
\label{eqn:L_D}
\mathcal{L}_{D} =  -\mathbb{E}_{\bm{x} \thicksim P_{r}}[\mathrm{log} D(\bm{x})] - \mathbb{E}_{\bm{z} \thicksim P_{\bm{z}}}[\mathrm{log} (1 - D(G(\bm{z}))],
\end{equation}
while network $G$ tries to minimize
\begin{displaymath}
\label{eqn:L_GD_traiditional}
\mathcal{L}_{GD}' = -\mathbb{E}_{\bm{z} \thicksim P_{\bm{z}}}[\log{D(G(\bm{z}))}].
\end{displaymath}

In practice, the distributions of ``real'' and ``fake'' images may not overlap with each other, especially at the early stage of the training process. Hence, the discriminative network $D$ can separate them perfectly. That is, we always have $D(\bm{x})\to1$ and $D(\bm{x}')\to0$, where $\bm{x}'=G(\bm{z})$ is the generated image. Therefore, when updating network $G$, the gradient $\partial \mathcal{L}_{GD}'/\partial D(\bm{x}') \to -\infty$. So the training process of network $G$ will be unstable. Recent works~\cite{arjovsky2017towards,arjovsky2017wasserstein} also theoretically show that training GAN often has to deal with the unstable gradient of $G$.

To address this problem, we propose using a mean feature matching objective for the generator. The objective requires the center of the features of the synthesized samples to match the center of the features of the real samples. Let $f_{D}(\bm{x})$ denote features on an intermediate layer of the discriminator, $G$ then tries to minimize the loss function
\begin{equation}
\label{eqn:L_GD_stat}
\mathcal{L}_{GD} =  \frac{1}{2}||\mathbb{E}_{\bm{x} \thicksim P_{r}}f_{D}(\bm{x}) - \mathbb{E}_{\bm{z} \thicksim P_{\bm{z}}}f_{D}(G(\bm{z}))||_2^2.
\end{equation}
In our experiment, for simplicity, we choose the input of the last Fully Connected (FC) layer of network $D$ as the feature $f_D$. Combining the features of multiple layers could marginally improve the converging speed. In the training stage, we estimate the mean feature using the data in a minibatch. We also use moving historical averages to make it more stable.

Therefore, in the training stage, we update network $D$ using Eq.~\ref{eqn:L_D}, and update network $G$ using Eq.~\ref{eqn:L_GD_stat}. Using this asymmetrical loss for training GAN has the following three advantages: 1) since Eq.~\ref{eqn:L_GD_stat} increases with the separability, the $\ell_2$ loss on the feature center solves the gradient vanishing problem; 2) when the generated images are good enough, the mean feature matching loss becomes zero, making the training more stable; 3) compared with WGAN~\cite{arjovsky2017wasserstein}, we do not need to clip the parameters. The discriminative power of network $D$ can be kept.

\subsection{Mean Feature Matching for Conditional Image Generation}
\label{sec:mean_feature_matching_based CGAN}
In this section, we introduce mean feature matching for conditional image generation. Supposing we have a set of data belonging to $K$ categories, we use the network $C$ to measure whether an image belongs to a specific fine-grained category. Here we use a standard method for classification. The network $C$ takes in $\bm{x}$ as input and outputs a $K$-dimensional vector,
which then turns into class probabilities using a softmax function. The output of each entry represents the posterior probability $P(c|\bm{x})$. In the training stage, the network $C$ tries to minimize the softmax loss

\begin{equation}
\label{eqn:L_C}
\mathcal{L}_{C} =  -\mathbb{E}_{\bm{x} \thicksim P_{r}}[\log{P(c|\bm{x})}].
\end{equation}

\noindent For the network $G$, if we still use the similar softmax loss function as in Eqn.~\ref{eqn:L_C}, it will suffer from the same gradient instability problem as described in ~\cite{arjovsky2017towards}.

Therefore, we propose using the mean feature matching objective for generative network $G$. Let $f_{C}(\bm{x})$ denote features on an intermediate layer of the classification, then $G$ tries to minimize:
\begin{equation}
\label{eqn:L_GC_stat}
\mathcal{L}_{GC} =  \frac{1}{2}\sum_{c}||\mathbb{E}_{\bm{x} \thicksim P_{r}}f_{C}(\bm{x}) - \mathbb{E}_{\bm{z} \thicksim P_{\bm{z}}}f_{C}(G(\bm{z}, c))||_2^2.
\end{equation}
Here, we choose the input of the last FC layer of network $C$ as the feature for simplicity. We also try to combine features of multiple layers, it only marginally improves the ability to preserve the identity of network $G$. Since there are only a few samples belonging to the same category in a minibatch, it is necessary to use moving averages of features for both real and generated samples.

\begin{figure*}[t]
  \centering
  \includegraphics[width=1.9\columnwidth]{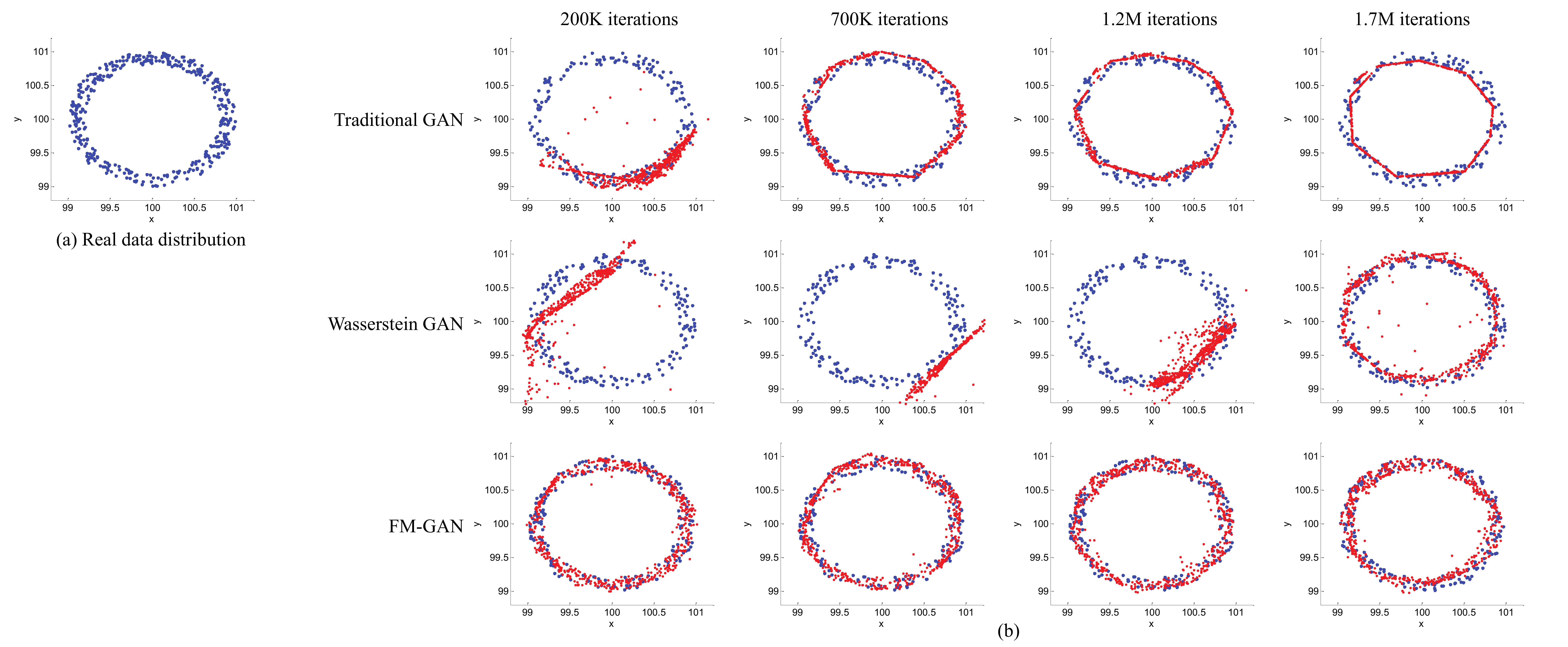}\\
  \caption{Results on a toy example for different generative models. The blue dots are the real points, the red dots are the generated points. a) The real data distribution which is like a ``ring". b) The generated points
  by traditional GAN, WGAN and mean feature matching GAN at different iterations.}\label{fig:toy_example}
  \vspace{-0.5cm}
\end{figure*}

\subsection{Pairwise Feature Matching}
\label{sec:pairwise_feature_matching_based_CVAEGAN}
Although, using mean feature matching could prevent all outputs from moving toward a single point, thus reducing the likelihood of mode collapse, it does not completely solve this problem. Once mode collapse occurs, the generative network outputs the same images for different latent vectors, thus the gradient descent will not be able to separate these identical outputs. Moreover, despite the generated samples and real samples having the same feature center, they may have different distributions.

In order to generate diverse samples, DCGAN~\cite{radford2015unsupervised} uses Batch Normalization, McGAN~\cite{mroueh2017mcgan} uses both mean and covariance  feature statistics, and Salimans \etal~\cite{salimans2016improved} use minibatch discrimination. They are all based on using multiple generated examples. Different from these methods, we add an encoder network $E$ to obtain a mapping from the real image $\bm{x}$
to the latent space $\bm{z}$. Therefore, we explicitly set up the relationship between the latent space and real image space.

Similar to VAE, for each sample, the encoder network outputs the mean and covariance of the latent vector, i.e., $\mu$ and $\epsilon$. We use the $KL$ loss to reduces the gap between the prior $P(\bm{z})$ and the proposal distributions, i.e.,

\begin{equation}
\mathcal{L}_{KL} = \frac{1}{2}\left(\bm{\mu}^T\bm{\mu}+sum(\exp (\bm{\epsilon})-\bm{\epsilon}-1)\right).
\end{equation}

We can then sample the latent vector $\bm{z} = \mu+\bm{r} \odot exp(\epsilon)$, where $\bm{r} \thicksim N(\bm{0}, \bm{I})$ is a random vector and $\odot$ represents the element-wise multiplication. After obtaining a
mapping from $\bm{x}$ to $\bm{z}$, we obtain the generated image $\bm{x}'$ with network $G$. Then, we add a $\ell_2$ reconstruction loss and pair-wise feature matching loss between $x$ and $x'$,

\vspace{-0.5cm}
\begin{equation}
\begin{split}
\label{eqn:L_G}
\mathcal{L}_{G} = \frac{1}{2}(||\bm{x}- \bm{x}'||_2^2&+||f_D(\bm{x}) - f_D(\bm{x}')||_2^2\\
&+||f_C(\bm{x}) - f_C(\bm{x}')||_2^2),
\end{split}
\end{equation}

where, $f_D$ and $f_C$ are the features of an intermediate layer of discriminative network $D$ and classification network $C$, respectively.

\subsection{Objective of CVAE-GAN}
\label{sec:objective_of_CVAE_GAN}
The goal of our approach is to minimize the following loss function:

\vspace{-0.5cm}
\begin{equation}
\label{eqn:L_CVAE-GAN}
\mathcal{L} = \mathcal{L}_{D} + \mathcal{L}_{C} + \lambda_1 \mathcal{L}_{KL} + \lambda_2 \mathcal{L}_{G} + \lambda_3 \mathcal{L}_{GD} + \lambda_4\mathcal{L}_{GC},
\end{equation}

where the exact forms of each of the terms are presented in Eqns.~\ref{eqn:L_D}$\thicksim$\ref{eqn:L_G}. Every term of the above formula is meaningful. $\mathcal{L}_{KL}$ is only related to the encoder network $E$. It represents whether the distribution of the latent vector is under expectation. $\mathcal{L}_{G}$, $\mathcal{L}_{GD}$ and $\mathcal{L}_{GC}$ are related to the generative network $G$. They represent whether the synthesized image is similar to the input training sample, the real image, and other samples within the same category, respectively. $\mathcal{L}_{C}$ is related to the classification network $C$, which represents the capability of the network to classify images from different categories, and $\mathcal{L}_{D}$ is related to the discriminative network, which represents how good the network is at distinguishing between real/synthesized images. All these objectives are complementary to each other, and ultimately enable our algorithm to obtain superior results. The whole training procedure is described in Algorithm~\ref{alg:pipeline}. In our experiments. we empirically set $\lambda_1 = 3$, $\lambda_2 = 1$, $\lambda_3 = 10^{-3}$ and $\lambda_4 = 10^{-3}$.

\section{Analysis of Toy Example}
\label{sec:analysis_with a toy example}
In this section, we present and demonstrate the benefits of the mean feature matching based GAN with a toy example. We assume that we have a real data distribution which is a ``ring" as shown in Figure~\ref{fig:toy_example}(a). The center of the ring is set to $(100, 100)$, such that it is far from the generated distribution at the beginning.
We compare the traditional GAN, WGAN, and the mean feature matching based GAN introduced in Section~\ref{sec:mean_feature_matching_based GAN} to learn the real data distribution.

The three compared models share the same settings. Generator $G$ is an MLP with 3 hidden layers with 32, 64, and 64 units, respectively. Discriminator $D$ is also an MLP with 3 hidden layers with 32, 64, and 64 units, respectively.
We use RMSProp and a fixed learning rate of $0.00005$ for all methods. We trained each model for $2M$ iterations until they all converge. The generated samples of each model at different iterations
 are plotted in Figure~\ref{fig:toy_example}. From the results we can observe that: 1) For traditional GAN (first row in Figure~\ref{fig:toy_example}(b)), the generated samples only lie in a limited area of the real data distribution, which is known as the mode collapse problem. This problem always exists during the training process. 2) For WGAN (second row in Figure~\ref{fig:toy_example}(b)), it cannot learn the real data distribution at early iterations, we think this problem is caused by the clamping weights trick, which influence $D$'s ability to distinguishing between real and fake samples. We also tried to vary the clamp values to accelerate the training process, and find that if the value is too small, it will cause a gradient vanishing problem. If too large, the network will diverge. 3) The third row shows the results of the proposed feature matching based GAN. It correctly learns the real data distribution the fastest.

\begin{algorithm}[t] \small
\caption{The training pipeline of the proposed CVAE-GAN algorithm.}
\label{alg:pipeline}

\begin{algorithmic}
\REQUIRE $m$, the batch size. $n$, class number. $\theta_{E}$, initial $E$ network parameters. $\theta_{G}$, initial $G$ network parameters. $\theta_{D}$, initial $D$ network parameters. $\theta_{C}$, initial $C$ network parameters, $\lambda_1 = 3$, $\lambda_2 = 1$, $\lambda_3 = 10^{-3}$ and $\lambda_4 = 10^{-3}$.
\end{algorithmic}

\begin{algorithmic}[1]
\WHILE{$\theta_{G}$ has not converged}
\STATE Sample $\{x_{r}, c_{r}\} \sim P_{r}$ a batch from the real data;
\STATE $\mathcal{L}_{C}$ $\leftarrow$ $-$log($P(c_{r}|x_{r})$)
\STATE $z$ $\leftarrow$ $E(x_{r}, c_{r})$
\STATE $\mathcal{L}_{KL}$ $\leftarrow$ $KL(q(z|x_{r}, c_{r})||P_{\bm{z}})$
\STATE $x_{f}$ $\leftarrow$ $G(z, c_{r})$
\STATE Sample $z_{p} \sim P_{\bm{z}}$ a batch of random noise, sample $c_{p}$ a batch of random classes;
\STATE $x_{p}$ $\leftarrow$ $G(z_{p}, c_{p})$
\STATE $\mathcal{L}_{D}$ $\leftarrow$ $-$(log($D(x_{r})$) + log(1- $D(x_{f})$) + log(1 - $D(x_{p})$))
\STATE Calculate $x_{r}$ feature center $\frac{1}{m}\sum_{i}^{m}f_{D}(x_{r})$ and $x_{p}$ feature center $\frac{1}{m}\sum_{i}^{m}f_{D}(x_{p})$;
\STATE $\mathcal{L}_{GD}$ $\leftarrow$ $ \frac{1}{2}||\frac{1}{m}\sum_{i}^{m}f_{D}(x_{r}) - \frac{1}{m}\sum_{i}^{m}f_{D}(x_{p})||_2^2$
\STATE Calculate each class $c_{i}$ feature center $f_{C}^{c_{i}}(x_{r})$ for $x_{r}$ and $f_{C}^{c_{i}}(x_{p})$ for $x_{p}$ using moving average method;
\STATE $\mathcal{L}_{GC}$ $\leftarrow$ $ \frac{1}{2}\sum_{c_{i}}||f_{C}^{c_{i}}(x_{r}) - f_{C}^{c_{i}}(x_{p})||_2^2$
\STATE $\mathcal{L}_{G}$ $\leftarrow$ $ \frac{1}{2}(||x_{r}- x_{f}||_2^2 + ||f_D(x_{r}) - f_D(x_{f})||_2^2 +||f_C(x_{r}) - f_C(x_{f})||_2^2)$

\STATE $\theta_{C}$ $\stackrel{+}{\longleftarrow}$ $-\nabla_{\theta_{C}}(\mathcal{L}_{C})$
\STATE $\theta_{D}$ $\stackrel{+}{\longleftarrow}$ $-\nabla_{\theta_{D}}(\mathcal{L}_{D})$
\STATE $\theta_{G}$ $\stackrel{+}{\longleftarrow}$ $-\nabla_{\theta_{G}}(\lambda_2\mathcal{L}_{G} + \lambda_3\mathcal{L}_{GD} + \lambda_4\mathcal{L}_{GC})$
\STATE $\theta_{E}$ $\stackrel{+}{\longleftarrow}$ $-\nabla_{\theta_{E}}(\lambda_1 \mathcal{L}_{KL} + \lambda_2\mathcal{L}_{G})$
\ENDWHILE
\end{algorithmic}
\end{algorithm}

\begin{figure*}[t]
  \centering
  \includegraphics[width=2\columnwidth]{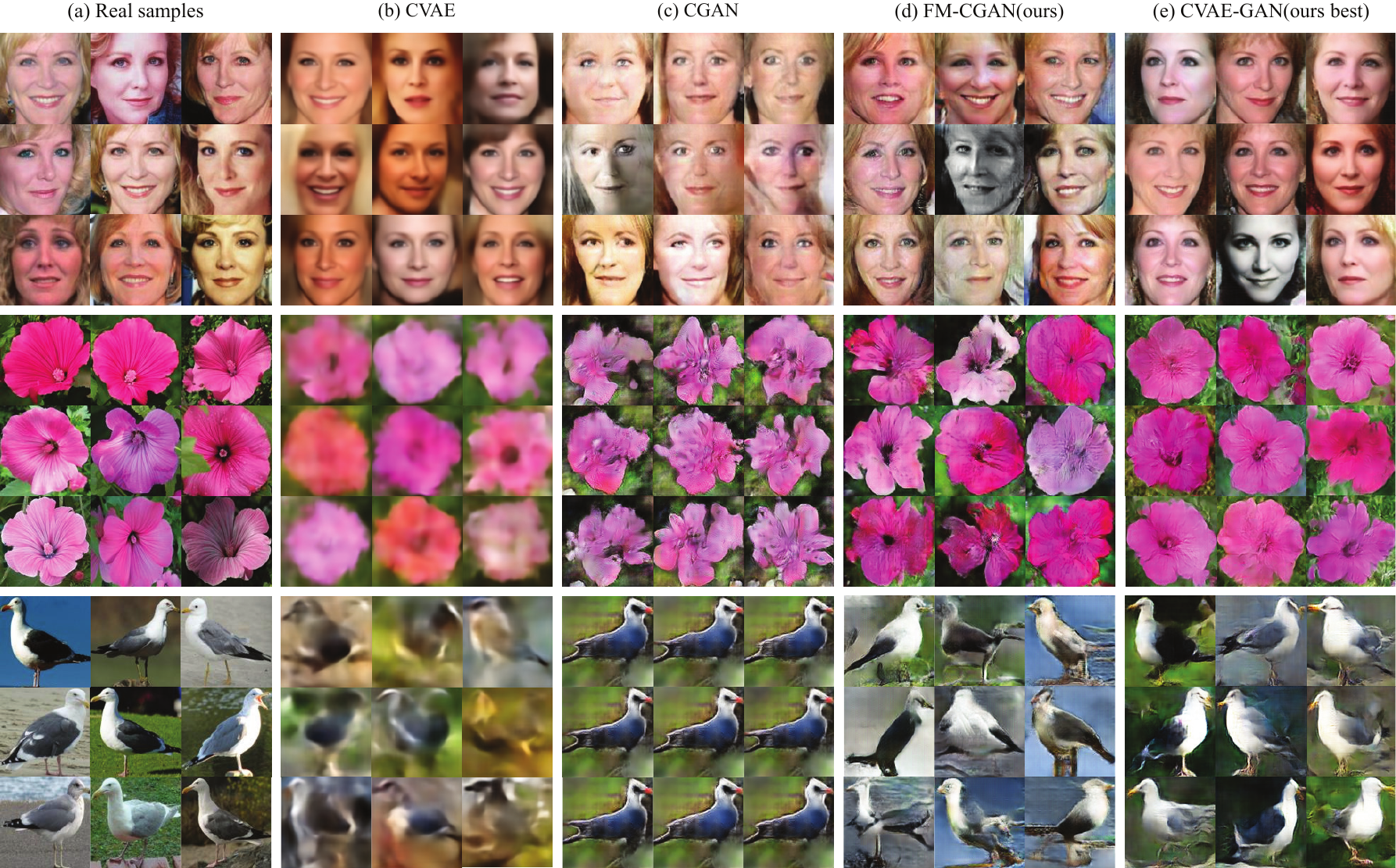}\\
  \caption{Comparison of randomly generated samples from different methods on FaceScrub~\cite{ng2014data}, 102 Category Flower datasets~\cite{nilsback2008automated} and CUB-200~\cite{WelinderEtal2010} datasets. a) 9 random real images from one category. b) Results from CVAE, which is blurry and cannot preserve the category identity, c) Results from traditional CGAN, it loses diversity and structure info. d) Results from our mean feature matching CGAN, showing diverse results, but also lose of structure info. e) Results from our CVAE-GAN, which shows realistic, diversity and category-keeping results.}\label{fig:mnist_cvae_cgan_cvaegan}
  \label{fig:comparison_with_CVAE_CGAN}
  \vspace{-0.5cm}
\end{figure*}

\section{Experiments}
\label{sec:experiments}
In this section, we use experiments to validate the effectiveness of the proposed method. We evaluate our model on three datasets: the FaceScrub~\cite{ng2014data},the 102 Category Flower~\cite{nilsback2008automated}, and CUB-200~\cite{WelinderEtal2010} datasets. These three datasets contain three completely different objects, which are human faces, birds, and flowers, respectively.

The sizes of input and synthesized images are $128 \times 128$ for all experiments. For the FaceScrub dataset, we first detect the facial region with the JDA face detector~\cite{chen2014joint}, and then locate five facial landmarks (two eyes, nose tip and two mouth corners) with SDM~\cite{xiong2013supervised}. After that, we use similarity transformation based on the facial landmarks to align faces to a canonical position.  Finally, we crop a $128 \times 128$ face region centered around the nose tip. For the 102 Category Flower dataset, we tightly crop a rectangle region based on the ground-truth mask which contains the flower, and then resize it into $128 \times 128$. For the CUB-200 dataset, we just use the original images from the dataset.

In our experiments, the encoder network $E$ is a GoogleNet~\cite{szegedy2015going}, The category information and the image is merged at the last FC layer of the $E$ network. The $G$ network consists of 2 fully-connected layers, followed by 6 deconv layers with 2-by-2 upsampling. The convolution layers have 256, 256, 128, 92, 64 and 3 channels with filter size of 3 $\times$ 3, 3 $\times$ 3, 5 $\times$ 5, 5 $\times$ 5, 5 $\times$ 5, 5 $\times$ 5. For the $D$ network we use the same $D$ network as the DCGAN~\cite{radford2015unsupervised}.For the $C$ network, we use an Alexnet \cite{krizhevsky2012imagenet} structure, and change the input to $128 \times 128$. We fix the latent vector dimension to be 256 and find this configuration sufficient for generating images. The batch normalization layer~\cite{ioffe2015batch} is also applied after each convolution layer. The model is implemented using the deep learning toolbox Torch.

\subsection{Visualization comparison with other models}

In this experiment, we compare the proposed mean feature matching based CGAN introduced in Section~\ref{sec:mean_feature_matching_based CGAN} (FM-CGAN), and CVAE-GAN model with other generative models for image synthesis of fine-grained images.

In order to fairly compare each method, we use the same network structure and same training data for all methods. All networks are trained from scratch. In the testing stage, the network architectures are the same. All three methods only use network $G$ to generate images. Therefore, although our approach has more parameters in the training stage, we believe this comparison is fair.

We conduct experiments on three datasets: FaceScrub, 102 Category Flower and  CUB-200 dataset. We perform category conditioned image generation for all methods.  For each dataset, all methods are trained with all the data in that dataset.  In the test stage, we first randomly chose a category $c$, and then randomly generate samples of that category by sampling the latent vector $\bm{z} \thicksim N(\bm{0}, \bm{I})$. For evaluation, we visualize the samples generated from all methods.

The comparison results are presented in Figure~\ref{fig:comparison_with_CVAE_CGAN}. All images are randomly selected without any personal bias. We observe that images generated by CVAE are often blurry. For traditional CGAN, the variation within a category is very small, which is because of the mode collapse. For FM-CGAN, we observe clear images with well preserved identities, but some images lose the structure of an object, such as the shape of the face. On the other hand, images generated by the proposed CVAE-GAN models look realistic and clear, and are non-trivially different from each other, especially for view-point and background color. Our model is also able to keep the identity information. It shows the strength of the proposed CVAE-GAN method.

\subsection {Quantitative Comparison}
\label{sec:Quantitative_comparison}
Evaluating the quality of a synthesized image is challenging due to the variety of probabilistic criteria ~\cite{theis2015note}. We attempt to measure the generative model on three criteria: discriminability,
diversity and realism.

We use face images for this experiment. First, we randomly generate 53$k$ samples (100 for each class) from CVAE, CGAN, FM-CGAN, and CVAE-GAN models for evaluation.

To measure discriminability, we use a pre-trained face classification network on the real data. Here we use GoogleNet~\cite{szegedy2015going}. With this trained model, we evaluate the top-$1$ accuracy of the generated samples from each method. The results are shown in Table~\ref{tab:result_of_quantitative comparison}. Our model achieves the best top-$1$ accuracy with a big gap to other generative models. This demonstrates the effectiveness of the proposed method.

Following the method in~\cite{salimans2016improved}, we use the \emph{Inception Score} to evaluate the realism and diversity of generated samples. We train a classification model on the CASIA~\cite{yi2014learning} datasets, and adopt $exp(\mathbb{E}_{\bm{x}}KL(p(y|\bm{x})||p(y)))$ as the metric to measure the realism and diversity of the generative models, where $p(y|\bm{x})$ represents the posterior probability of each class of generated samples. Images that contain meaningful objects should have a conditional label distribution $p(y|\bm{x})$ with low entropy. Moreover, if the model generate diverse images, the marginal $p(y)=\int p(y|G(\bm{z}))d\bm{z}$ should have high entropy. A larger score means the generator can produce more realistic and diverse images. The results are shown in Table~\ref{tab:result_of_quantitative comparison}. Our proposed CVAE-GAN and FM-CGAN achieve better scores than the other models, which are also very close to the real data.

\begin{figure}[t]
  \centering
  \includegraphics[width=1\columnwidth]{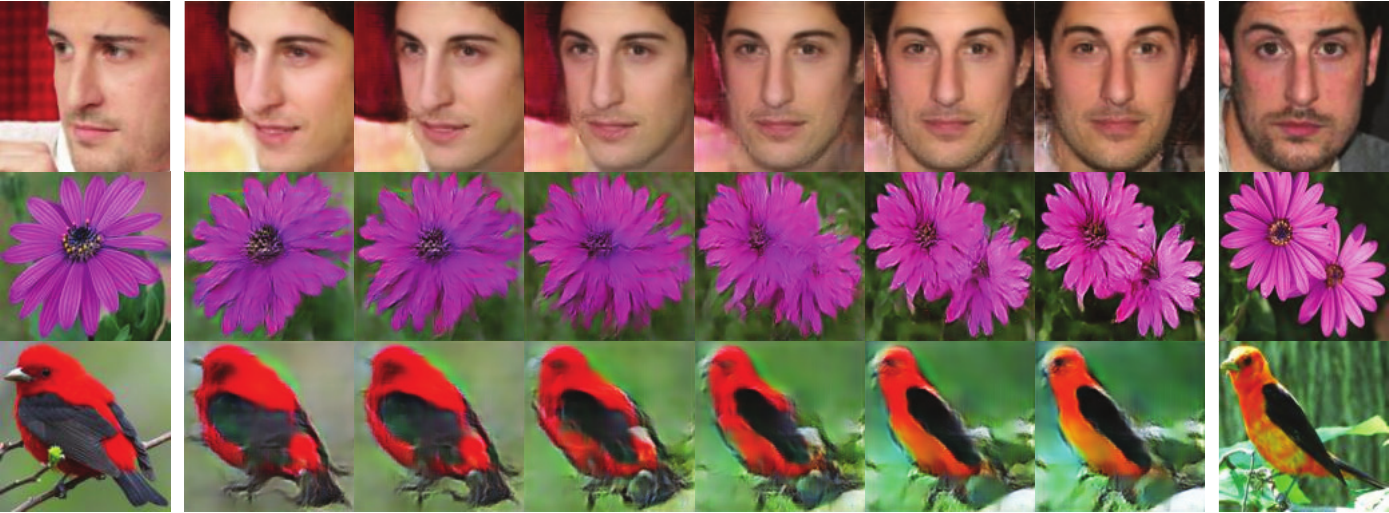}\\
  \caption{Results of attributes morphing.}\label{fig:attributes_morphing}
  \vspace{-0.3cm}
\end{figure}

\begin{table}[t] \small
\centering
\begin{tabular}{p{1.3cm}|p{0.8cm}|p{0.8cm}|p{0.8cm}|p{0.8cm}|p{0.8cm}}
    \hline
                   & Real data & CVAE & CGAN & FM-CGAN & CVAE-GAN \\
    \hline
    Top-1 acc      & 99.61\%   & 8.09\%     & 61.97\%      &  79.76\%       & 97.78\%    \\
    \hline
    Realism          & 20.85  & 10.29     & 15.79     &   19.40      & 19.03         \\
    \hline
\end{tabular}
\caption{Quantitative result of generated image quality of different methods. Please refer to Section~\ref{sec:Quantitative_comparison} for details.}
\label{tab:result_of_quantitative comparison}
  \vspace{-0.5cm}
\end{table}

\subsection{Attributes Morphing}
\label{sec:the_same_latent_vector_represent_same_attribute}

In this part, we validate that the attribute in the generated images will continuously change with the latent vector. We call this phenomenon attribute morphing. We also test our model on the FaceScrub, CUB-200 and 102 Category
Flower datasets. We first select a pair of images $\bm{x}_1$ and $\bm{x}_2$ in the same category, and then extract the latent vector $\bm{z}_1$ and $\bm{z}_2$ using the encoder network $E$. Finally, we obtain a series of latent vectors $\bm{z}$ by linear interpolation,i.e., $\bm{z} = \alpha \bm{z}_1 + (1 - \alpha)\bm{z}_2, \alpha \in [0, 1]$.
Figure~\ref{fig:attributes_morphing} shows the results of attribute morphing. In each row, the attribute, such as pose, emotion, color, or flower number, gradually changes from left to right.

\subsection{Image Inpainting}
\begin{figure}[t]
  \centering
  \includegraphics[width=1\columnwidth]{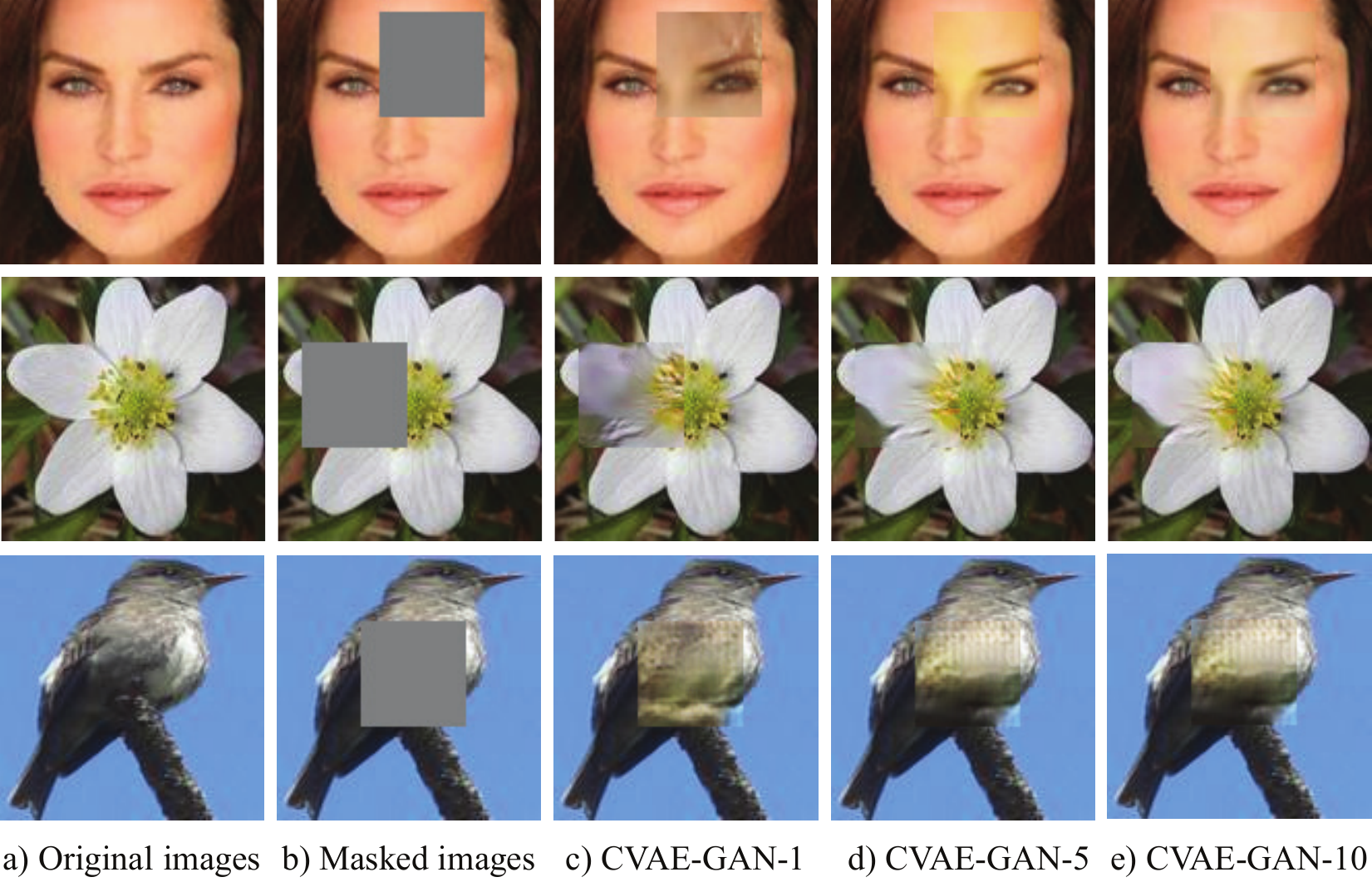}\\
  \caption{Result of image inpainting using our proposed model CVAE-GAN-$1\sim 10$ shows the results of  iteration $1\sim 10$.}\label{fig:inpainting result}
  \vspace{-0.5cm}
\end{figure}

In this part, we show that our model can also be applied to image inpainting. We first randomly corrupt a
$50 \times 50$ patch of an original $128 \times 128$ image $\bm{x}$ (Fig.\ref{fig:inpainting result}b), and then feed it to the $E$ network to obtain a latent vector $\bm{z}$, then we can synthesize an image $\bm{x}'$ by $G(\bm{z}, c)$ where $c$ is the class label,
then we update the image by the following equation,i.e.,

\vspace*{-0.3cm}
\begin{equation}
\label{eqn:inpainting update}
\bm{x} = M \odot \bm{x}' + (1 - M) \odot \bm{x},
\vspace*{-0.3cm}
\end{equation}

where $M$ is the binary mask for the corrupted patch, and $\odot$ denotes the element-wise product. So $(1-M)\odot \bm{x}$ is the uncorrupted area in the original image. The inpainting results are shown in
Figure~\ref{fig:inpainting result} (c). We should emphasize that all input images are downloaded from websites, with none of them belonging to the training data. We can iteratively feed the resulting images into the model to obtain a better results, as shown in Figure~\ref{fig:inpainting result} (d,e).

\subsection{Comparing Different Combination of Losses}

In our model, we propose using pairwise feature matching at the image pixel level, the feature level in the classification network $C$ and the discriminative network $D$ to update the network $G$. To understand the effects of each loss component, we separate the $\mathcal{L}_{G} + \mathcal{L}_{GD} + \mathcal{L}_{GC}$ to three parts: $\mathcal{L}_{G}(img) + \mathcal{L}_{G}(D) + \mathcal{L}_{G}(C)$, where $\mathcal{L}_{G}(img)$ is the $\ell_2$ distance at the pixel level of the image, $\mathcal{L}_{G}(D)$ is the $\ell_2$ distance at the feature level in the discriminative network $D$, $\mathcal{L}_{G}(C)$ is the $\ell_2$ distance at the feature level in the classification network $C$.

\begin{figure}[!tp]
  \centering
  \includegraphics[width=1\columnwidth]{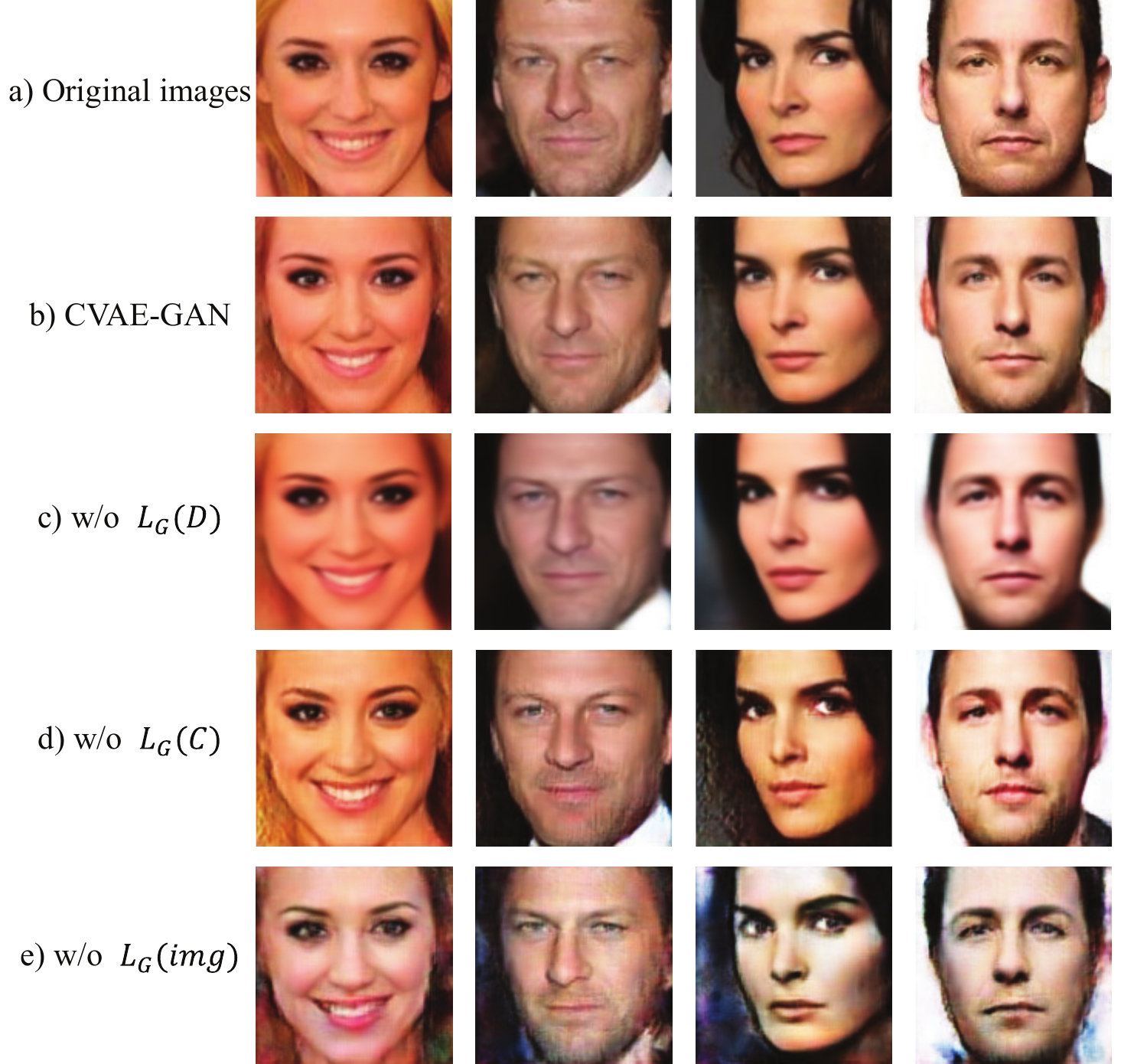}\\
  \caption{Visualization comparison between different generator $G$, each trained with different combination of losses.}
  \label{fig:loss_combination_results}
  \vspace{-0.2cm}
\end{figure}

We repeat the training of the CVAE-GAN model with the same settings but using different combination of losses in $\mathcal{L}_{G}(img)$, $\mathcal{L}_{G}(D)$, and $\mathcal{L}_{G}(C)$, and compared the quality of the reconstructed samples. As shown in Fig.~\ref{fig:loss_combination_results}, we find that removing the adversarial loss $\mathcal{L}_{G}(D)$ will cause the model to generate blurry images. Removing the pixel level reconstruction loss $\mathcal{L}_{G}(img)$ causes images to lose details. Lastly, if we remove the feature level loss $\mathcal{L}_{G}(C)$ in the classification network $C$, the generated samples will lose category info. Despite this, our model produces best results.

\subsection{CVAE-GAN for Data Augmentation}
\label{sec:CVAE-GAN_for_data_augmentation}

We further show that the images synthesized from our model can be used for data augmentation for training better face recognition model. We use the FaceScrub dataset as training data, and test using the LFW~\cite{learned2016labeled} dataset.

We experiment with two data augmentation strategies: 1) generating more images for existing identities in the training datasets; 2) generating new identities by mixing different identities. We
test these two kinds of data augmentation methods. For 1), we randomly generate about 200 images per person, totaling 100$k$ images.
For 2), we create 5$k$ new identities by randomly mixing the label of three different existing identities, and generate 100 images for each new identity. For both strategies, the generated images are combined with the Facescrub dataset to train a face recognition model.

In the testing stage, we directly use the cosine similarity of features to measure the similarity between two faces. In Table~\ref{tab:result_of_data_augment}, we compare face verification accuracy
on the LFW dataset with and without additional synthesized faces. With the data augmentation of new identities, we achieve about $1.0\%$ improvement in accuracy compared with no augmentation. This demonstrates
that our generative network has a certain extrapolation ability.

\begin{table}[t]\small
  \centering
\begin{tabular}{l|c|c}
  \hline
  Method & Training Data & Accuracy \\
  \hline
  no data augmentation & $80K$ & $91.87\%$ \\
  existing identities augmentation & $80K + 100K$& $92.77\%$ \\
  $5k$ new identities augmentation & $80K + 500K$& $92.98\%$ \\
  \hline
\end{tabular}
\caption{Results of face data augmentation.}
\label{tab:result_of_data_augment}
\vspace{-0.5cm}
\end{table}

\section{Conclusion}
In this paper, we propose a new CVAE-GAN model for fine-grained category image generation. The superior performance on three different datasets demonstrates the ability to generate various kinds of objects. The proposed method can
support a wide variety of applications, including image generation, attribute morphing, image inpainting, and data augmentation for training better face recognition models. Our future work will explore how to generate
samples of an unknown category, such as face images of a person that do not exist in the training dataset.

\clearpage
{\small
\bibliographystyle{ieee}
\bibliography{egbib}
}

\end{document}